%% file: main.tex
\documentclass[10pt,twocolumn,letterpaper]{article}

\usepackage{cvpr}              

\input{preamble}

\definecolor{cvprblue}{rgb}{0.21,0.49,0.74}
\usepackage[pagebackref,breaklinks,colorlinks,citecolor=cvprblue]{hyperref}
\usepackage[accsupp]{axessibility} 

\def\paperID{2990} 
\def\confName{CVPR}
\def\confYear{2024}

\title{Unsegment Anything by Simulating Deformation}

\author{Jiahao Lu\qquad Xingyi Yang\qquad Xinchao Wang\thanks{Corresponding Author.}\\
National University of Singapore\\
{\tt\small \{jiahao.lu, xyang\}@u.nus.edu, xinchao@nus.edu.sg} \\
}

\begin{document}
\maketitle
\begin{abstract}
Foundation segmentation models, while powerful, pose a significant risk: they enable users to effortlessly extract any objects from any digital content with a single click, potentially leading to copyright infringement or malicious misuse. To mitigate this risk, we introduce a new task ``Anything Unsegmentable'' to grant any image  ``the right to be unsegmented''. The ambitious pursuit of the task is to achieve highly transferable adversarial attack against all prompt-based segmentation models, regardless of model parameterizations and prompts. We highlight the non-transferable and heterogeneous nature of prompt-specific adversarial noises. Our approach focuses on disrupting image encoder features to achieve prompt-agnostic attacks. Intriguingly, targeted feature attacks exhibit better transferability compared to untargeted ones, suggesting the optimal update direction aligns with the image manifold. Based on the observations, we design a novel attack named Unsegment Anything by Simulating Deformation (UAD). Our attack optimizes a differentiable deformation function to create a target deformed image, which alters structural information while preserving achievable feature distance by adversarial example. Extensive experiments verify the effectiveness of our approach, compromising a variety of promptable segmentation models with different architectures and prompt interfaces. We release the code at \href{https://github.com/jiahaolu97/anything-unsegmentable}{https://github.com/jiahaolu97/anything-unsegmentable}.
\end{abstract}

\section{Introduction}
\label{sec:intro}


The emergence of promptable segmentation models, exemplified by the Segment Anything Model (SAM) \cite{kirillov2023segment}, has demonstrated astonishing generalization capabilities across unseen data distributions and downstream tasks. Nevertheless, while these models offer remarkable convenience, they also introduce potential risks. They enable covert and effortless content filching, allowing unauthorized users to segment and misappropriate visual content with a single click. This is particularly concerning for artworks, digital designs, or promotional images, as segmenting such content can lead to commercial disputes. On the other hand, combining promptable segmentation models with generative AI techniques empowers users to perform precise in-place image editing or even 3D generation with a high level of realism \cite{yu2023inpaint, gao2023editanything, Qiuhong2023anything}. Segmentations taken out of their original context can be deceptive and may be misused, leading to unauthorized advertising and the generation of misleading news content, thereby posing potential societal risks.

The driving force behind this work is the need to address the above emerging risks with a proactive technical solution. We introduce an innovative task called ``Anything Unsegmentable'', which aimed at enhancing image resistance to any promptable segmentation model, consequently thwarting any unlawful attempts at image appropriation or manipulation. In pursuit of this ambitious objective, we propose a new adversarial attack Unsegment Anything by Simulating Deformation(UAD) emphasizing its remarkable transferability, as the adversarial perturbations remain model-agnostic and prompt-agnostic. This means that they can effectively confound segmentation foundational models, irrespective of their specific parameterizations and prompt formats.

While existing research has explored adversarial attacks targeting segmentation models \cite{xie2017adversarial, hendrik2017universal, arnab2018robustness, chen2022semantically}, our problem presents unique challenges. Firstly, it deviates from the existing approaches due to fundamental differences in input and output spaces. Semantic and panoptic segmentation models take images as input, producing pixel-level classification results. Conversely, SFMs generate binary masks in response to prompts, which can be either spatial (points, boxes, strokes) or semantic (speech, text, or exemplar references). Due to difference in input and output spaces, we need to devise novel attack objectives instead of encouraging pixel-level misclassifications. Secondly, recent studies \cite{huang2023robustness, qiao2023robustness, wang2023empirical} have demonstrated impressive robustness against various corruptions, surpassing the capabilities of ordinary segmentation models. Crafting attacks that can effectively transfer across these already robust foundation models poses a considerable challenge.

\begin{figure*}[t!]
  \centering
   \includegraphics[width=0.8\linewidth]{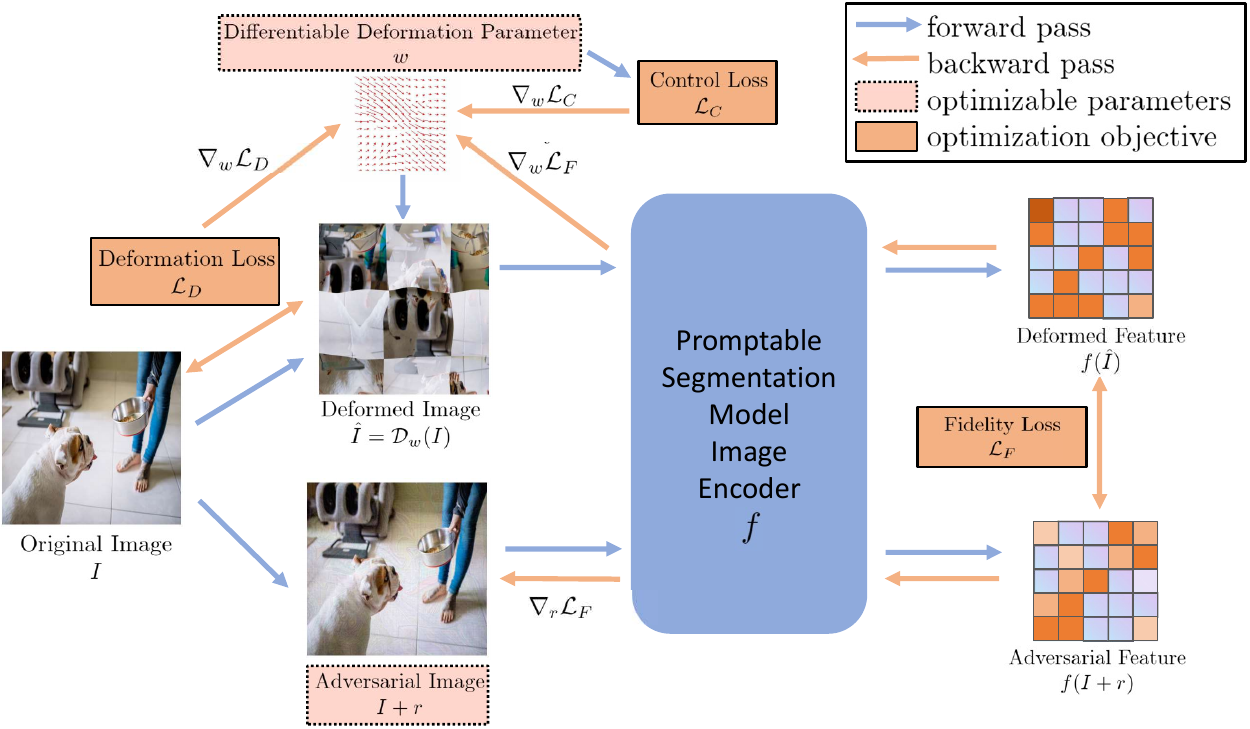}
   \caption{Pipeline of our attack. We optimize a deformation transform as well as the adversarial perturbation, to misguide the promptable segmentation model towards the deformed image.}
   \label{fig:pipeline}
\end{figure*}

We intend to present several key findings in this study. Firstly, through an examination of a prompt-specific adversarial perturbation algorithm proposed by concurrent research \cite{zhang2023attacksam}, we observed that adversarial noises derived from specific prompts typically exhibit high variance and lack generalizability across each other. In simpler terms, adversarial noise optimized for a specific prompt tends to overfit to that prompt and does not readily extend to other prompts. Secondly, we reveal that it is better to perturb features along the image manifold than against it to create a transferable adversarial sample. 
The adversarial noise crafted to shift away from original features in source model may be meaningless to target model and cannot arouse significant feature change, thus limiting their transferability. In contrast, targeted feature attacks bring similar feature disturbance in source and target models.
Lastly, we introduce a novel attack UAD, which optimizes an image deformation function as well as the adversarial perturbations. With our approach, the adversarial perturbation introduces shape misinformation, biasing segmentation results towards that particular deformation. Since the deformed image retains some natural image structure, such as textures and object parts, albeit distorted in shape, the feature distortion can be well transferred across segmentation models. This approach allows us to achieve prompt-agnostic and model-agnostic attacks. Empirical results demonstrate the superior effectiveness and transferability of our method compared to concurrent work and prior methods.

In brief, the contribution of this work is three-fold:
\begin{enumerate}
    \item We introduce a new challenging task \textit{Anything Unsegmentable}, which aims at prompt-agnostic and highly transferable adversarial attacks. 
    \item We reveal interesting findings on the robustness of promptable segmentation models, including (1) the overfitting nature of prompt-specific attacks and (2) targeted feature disruptions are more transferable than untargeted ones. The findings somehow disagree with previous observations on semantic segmentation models or classifiers, indicating the essential differences of their feature space. 
    \item We propose a new adversarial attack method \textit{UAD} as a progressive attempt to address the \textit{Anything Unsegmentable} task. We utilize differentiable deformation parameters to get an optimal target deformed image, which possess considerable structural change as well as feasible feature distance for adversarial updates. Compared with existing and contemporary works, our approach achieves state-of-the-art results, showing the effectiveness of our method. 
\end{enumerate}

\section{On the Robustness of Foundation Segmentation Models}
\subsection{Objective of Anything Unsegmentable Task}
For a promptable segmentation model, it typically contains an image encoder $f_{\theta^I}$, a prompt encoder $h_{\theta^P}$ and a mask decoder $g_{\theta^M}$.  The promptable segmentation task is designed to return a valid binary mask $M$ given an image $I$ and a prompt $P$:
\begin{equation}
    M = g_{\theta^M}(f_{\theta^I}(I), h_{\theta^P}(P)).
\end{equation}

The prompt $P$ offers high flexibility, encompassing spatial prompts such as foreground/background points, rough bounding boxes as well as semantic prompts that include high-level content descriptions like free-form text or memory prompts encapsulating prior segmentation information.

Our goal is to generate quasi-imperceptible noise $r$ to produce an adversarial image $I + r$ which significantly alters its segmentation outcome (e.g., yielding a low Intersection over Union (IoU)) regardless of the prompt applied.  Adversarial perturbation (or adversarial noise) $r$ is constrained in a feasible set, typically within an infinity norm ball with radius $\epsilon$, i.e. $ \|r\|_{\infty} \le \epsilon$. The Anything Unsegmentable task demands that the optimal adversarial perturbation $r^*$ is effective across various prompts and model parameters, formally represented as a solution to the subsequent optimization problem:

\vspace{-0.5em}
\begin{footnotesize}
\begin{equation}
\begin{aligned}
r^{*} &= \mathop{\arg\min}\limits_{\|r\|_{\infty} \le \epsilon} \mathop{\mathbb{E}}\limits_{\{\theta^I, \theta^P, \theta^M\}} \mathop{\mathbb{E}}\limits_{P} \quad IoU(M, M')\\
&= \mathop{\arg\min}\limits_{\|r\|_{\infty} \le \epsilon} \mathop{\mathbb{E}}\limits_{\{\theta^I, \theta^P, \theta^M\}} \mathop{\mathbb{E}}\limits_{P} \quad \\
& \quad IoU  \{ g_{\theta^M}(f_{\theta^I}(I), h_{\theta^P}(P)), g_{\theta^M}(f_{\theta^I}(I+r), h_{\theta^P}(P))\}.
\end{aligned}
\end{equation}
\end{footnotesize}
\vspace{-0.5em}

\subsection{Prompt-specific Attacks Transfer Poorly}
As outlined in the preceding section, the attacker's aim is to significantly alter the segmentation mask in response to any prompt. A straightforward idea is to craft an attack which deteriorates the segmentation outcome for a given prompt. This approach was recently employed in Attack-SAM \cite{zhang2023attacksam}. They introduced an innovative attack objective that minimizes the feature responses within the masked region. 

Their technique proved to be potent for the given prompt, however we found it challenging to generalize to alternative, unseen prompts. 
We offer visual evidence in Fig.\ref{fig:untransfer} and qualitative results in Tab. \ref{tab:mainexp} as evidence. \citet{zhang2023attacksam} admitted similar findings and  they introduced an improvement to enhance the transferability: instead of using only a single prompt, they randomly sample numerous point prompts, then execute the attack to invalidate the ensembled prompts.  While this improvement partially alleviates the issue, their subsequent work \cite{zheng2023pata} still demonstrates a noticeable performance gap between attacked prompts and unseen prompts. This observation highlights the challenge of prompt-based attacks being prone to overfit and lack generalizability. We further discuss the heterogenous and overfitting nature of prompt-specific attacks in Appendix Sec.\ref{sec:prompt-specific}.

\subsection{Perturbations Pointing Inside Image Manifold Transfer Better}

To avoid the overfitting behavior of prompt-specific attacks, we, therefore, endeavor to find an alternative approach for prompt-agnostic attacks. Considering image encoders have a more standardized and common functionality compared to prompt encoders or mask decoders \cite{yosinski2014transferable, hu2018deep, kornblith2019similarity, yang2022deep},
we opt to launch the attack from feature space of the image encoder.  

There existed many adversarial attack works launching attacks from feature space \cite{ganeshan2019fda, huang2019enhancing, zhou2018TAP, wang2021feature, wu2020adatt, Inkawhich2019AA, li2023ILPD}. Most of them fall into the category of \textit{untargeted feature disruption}, which maximize the distance between adversarial features and original features. The remaining fall under \textit{targeted feature perturbation} which brings adversarial sample closer to a specified input within the feature space. 
\vspace{-0.3cm}

\begin{figure}[h!]
  \begin{minipage}[t]{0.49\linewidth}
    \centering
    \includegraphics[width=\linewidth]{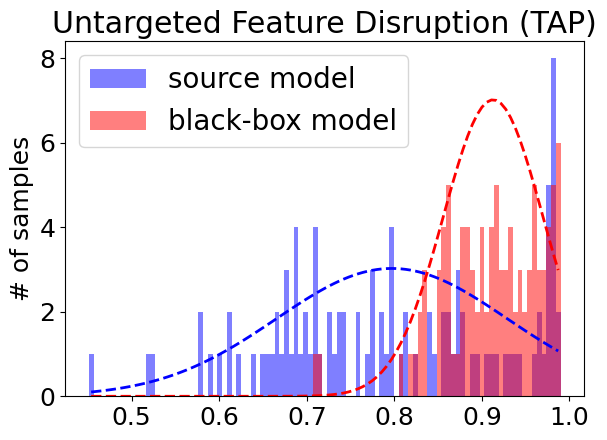}
    \label{fig:image1}
  \end{minipage}%
  \hfill
  \begin{minipage}[t]{0.49\linewidth}
    \centering
    \includegraphics[width=\linewidth]{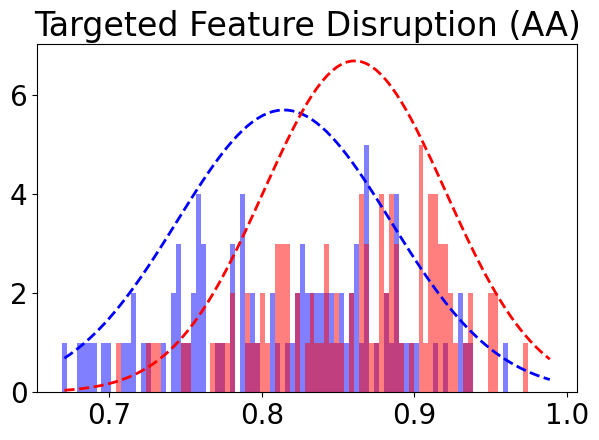}
    \label{fig:image2}
  \end{minipage}%
  \vspace{-2em}
  \caption{The histogram of feature similarities between adversarial and clean image, induced by  \textit{untargeted feature disruption} (left) and \textit{targeted feature disruption} (right) attacks on source (blue) and target (red) model.}
\label{fig:untarget}
\end{figure}

Previous literature has not provided clear evidence regarding the superiority of targeted or untargeted feature disruption in the context of classification or semantic segmentation. To our surprise, our investigation reveals that these two approaches exhibit distinct behaviors within the context of foundational segmentation models. We show in Fig.\ref{fig:untarget} that the feature perturbations resulted by untargeted attacks across models are significantly ineffective compared to targeted ones, even when the source and target models share similar architectural designs and training data. We selected TAP \cite{zhou2018TAP} as the untargeted feature disruption attack and AA \cite{Inkawhich2019AA} as a targeted attack. 
By evaluating feature similarity (cosine similarity of vectorized features of clean samples and adversarial samples) on the first 100 images in SAM-1B dataset, we found that targeted feature attacks can arouse similar level of feature disturbance on both models, while untargeted feature attacks can hardly arouse feature shifting on target model, suggesting the guidance from inside the image manifold is indispensable.

\begin{figure*}[t!]
  \centering
 \setlength{\abovecaptionskip}{0cm}
 \setlength{\belowcaptionskip}{-0.5cm}
   \includegraphics[width=\linewidth]{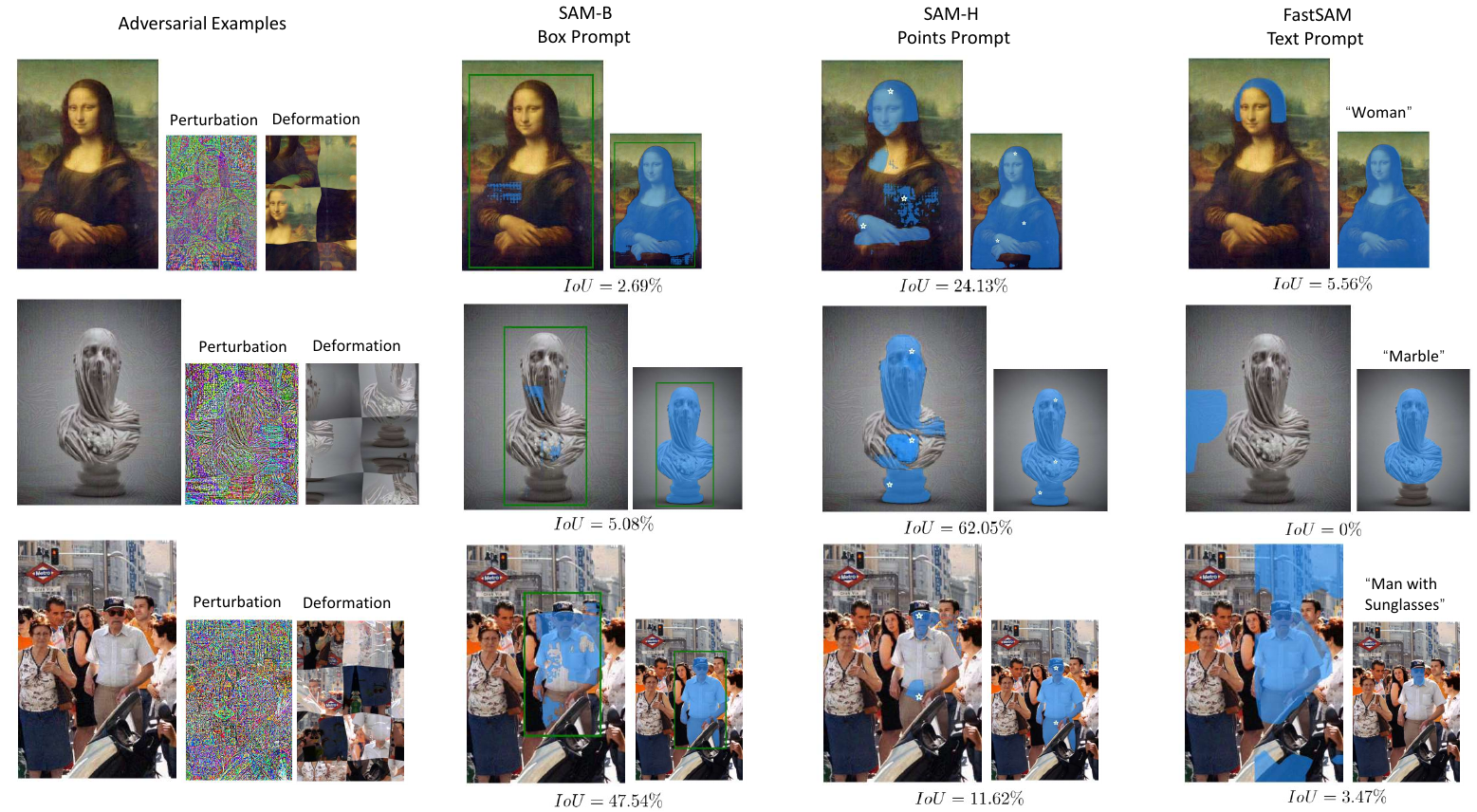}
   \caption{Adversarial examples crafted by our approach. In each row, we present, from left to right: the adversarial example, the adversarial perturbation, the optimized deformation target, attacked segmentation results and the original segmentation results on SAM-B (using a box prompt), SAM-H (using point prompts), and FastSAM (using a text prompt) respectively. The results demonstrate the high effectiveness of our approach against unseen models and versatile prompts.}
   \label{fig:examplars}
\end{figure*}



We attribute this surprising phenomenon to the inherent differences in tasks and their consequent effects on feature spaces. Classification or semantic segmentation models driven by class-discriminative objectives, tend to yield feature spaces rich in class-sensitive features. Consequently, any deviation from the original features introduces class-sensitive features from other classes, resulting in noticeable feature changes that lead to misclassifications. However in our case, the task doesn't involve category-specific information. Directions shifting away from the original features, may resemble meaningless random noise to target model and neglected by its robustness to image corruptions \cite{samrobust1}. In contrast, for in-distribution target images, adversarial perturbation towards them may effectively present a recognized and consistent update direction across all models. Briefly put, the adversarial perturbations pointing inside image manifold transfer better than perturbations pointing outside the image manifold for foundation segmentation models~\cite{li2023discrete}.

\section{Unsegment Anything by Simulating Deformation}
Building upon the insights from the preceding sections, it becomes evident that the target for feature perturbation should be close to natural images. Nevertheless, selecting a random target image from a population of images does not guarantee a sufficient level of structural dissimilarity with the original image, thus offering room for improvement.

We propose that optimization can be not limited to just the adversarial example, it should also encompass the target image. Our approach involves the optimization of a \textbf{differentiable image deformation function} applied to the original image to create the target for feature disruption. In essence, it drives the target to exhibit maximal shape deformation through optimization process. Our two-stage pipeline starts by identifying an optimal deformed image that balances high structural dissimilarity with closeness to natural image manifold and feasible set of adversarial samples. In the second stage, we align the features of adversarial sample to features of target deformed image.

\subsection{Stage One: Deformation}
We aim to make an ideal target deformed image through:
\begin{equation}
    \hat{I} = \mathcal{D}_w(I).
\end{equation}
where $\mathcal{D}_w$ is the deformation function to be optimized and $w$ is the differentiable deformation parameters which controls the deformation function. In practice, the design of $\mathcal{D}_w$ could be highly flexible. We can adopt any form of image transformation (e.g. rotation, translation, scaling, warping) to apply deformation as long as it has parameters to optimize. 

In our implementation, we use \textbf{flow fields} as $w$ to enable refined image deformation. The flow field $w_{ff}$ indicates a motion vector for each pixel position in the original image $I$. More specifically, for each pixel position $I^{(i, j)}$ in the original image, the direction of its motion is indicated by flow vector in corresponding position $w_{ff}^{(i, j)} = (\Delta u^i, \Delta v^j)$, and its destination position on deformed image is $\hat{I}^{(i + \Delta u^i, j + \Delta v^j)}$.  As the flow vector $ (\Delta u^i, \Delta v^j)$ could be fractional numbers and not necessarily integer, we use the differentiable bilinear interpolation \cite{jaderberg2015spatial} to transform input image with flow field.

\setParShrink
\subsubsection*{Deformation Loss}
The primary objective of deformation stage is to optimize a target image with maximal structural dissimilarity with original image, to misguide segmentation results. Any loss which encourages appearance deformations can serve the purpose, for example structural similarity index measure (SSIM) as the loss:

\begin{equation}
    \mathcal{L}_{D} = SSIM(\hat{I}, I).
    \vspace{3mm}
\end{equation}

In practice we found that SSIM is easy to achieve zero loss, causing deformation to stall.
To encourage greater deformation, we devised a strategy that combines patterns from various parts of the image. We concurrently optimize multiple flow fields, generating several image distortion results, and then combine them using pre-defined filter masks. Only a certain part of each deformation result can pass the filter mask and contribute to the final outcome, resulting a deformation that incorporates patches. This approach has proven more effective than just optimizing a single flow field, since the resulting patch contrastive patterns have even more significant structural differences than a single distorted target.

\subsubsection*{Control Loss}
One advantage of using a flow field lies in its flexibility, allowing us ample room to customize the regulation of the deformation function. For instance, we can apply Total Variation Loss to the flow field to promote locally smooth spatial transformations. Additionally, we can limit the variance of flow vectors to encourage globally uniform deformation like shifting. We employ a combination of variance loss and total variation loss to preserve local image patterns, creating the effect of assembling warped and shifted images:

\begin{equation}
\begin{aligned}
    &\mathcal{L}_{C} = \lambda_1 \mathcal{L}_{TV} + \lambda_2 \mathcal{L}_{var} \\
    &= \text{\resizebox{.8\hsize}{!}{$\lambda_1 \!\sum_{p}^\text{all pixels} \!\sum_{q \in \mathcal{N}(p)} \!\sqrt{\left\| \Delta u(p) - \Delta u(q) \right\|_2^2 + \left\| \Delta v(p) - \Delta v(q) \right\|_2^2}$}} \\
    & \text{\resizebox{.9\hsize}{!}{$+\!\lambda_2\!\sum_{p}^\text{all pixels}\! \sqrt{\! \|\Delta u(p) - \sum\limits_q^\text{all pixels}\!\frac{\Delta u(q)}{|q|}\|_2^2 + \|\Delta v(p) - \sum_q^\text{all pixels}\!\frac{\Delta v(q)}{|q|}\|_2^2}$}}.
\end{aligned}
\end{equation}

\subsubsection*{Fidelity Loss}
\vspace{-0.15em}
\setParRecover
As previously noted, not all solutions with low structural resemblance to original image are equally viable targets for adversarial noise to simulate.
Intuitively, we believe even though there are infinitely many $\hat{I}$ with low enough deformation loss $\mathcal{L}_D$, their feature distances to the feasible set of $I$ are not evenly distributed. Some targets among others are easier for the adversarial perturbation to approach, which results in better disturbance effect.

We take into the explicit consideration the difficulty for an adversarial example to approach the target deformed image in the feature space. To reduce the feature distance from the deformed target to the feasible set of adversarial examples, we introduce a `proxy' adversarial sample optimization process. This proxy sample $I_{proxy} = I + r'$ should be close to the boundary of feasible set, so that the distance from $\hat{I}$ to feasible set $\mathcal{N}^{adv}(I)$ can be approximated by the distance from $\hat{I}$ to $I + r'$. We impose a feature fidelity loss as negative cosine similarity between features to regulate the deformation: 

\begin{small}
\begin{equation}
\begin{aligned}
    &\mathcal{L}_F (\hat{I}, \mathcal{N}^{adv}(I)) \\
    &\approx \mathcal{L}_F (\hat{I}, I + r'^*)\\ 
    &= 1 - \frac{f_{\theta^I}(\hat{I}) \cdot f_{\theta^I}(I + r'^*)}{\|f_{\theta^I}(\hat{I})\|^2 \cdot \|f_{\theta^I}(I + r'^*)\|^2 }\\
    &\quad\text{where $r'^* = \mathop{\arg\min}\limits_{r'} \mathcal{L}_F (\hat{I}, I + r'^*)$}.
\end{aligned}
\label{eq:lf}
\end{equation}
\end{small}

As a conclusion, our desired target deformed image is an optimal solution to the following optimization problem:
\begin{equation}
    \hat{I}^* = \mathop{\arg\min}\limits_{\hat{I}} \mathcal{L}_D + \lambda_C \mathcal{L}_C + \lambda_F \mathcal{L}_F,
\end{equation}
where $\lambda_C$ and $\lambda_F$ are coefficients of each loss term. The optimization can be practically solved by gradient descent on differentiable deformation parameters $w$.

\subsection{Stage Two: Feature Simulation}
Once we have acquired the optimal target deformed image, the subsequent step aligns with those previous feature perturbation works, to encourage the adversarial perturbation close to a target image. We use the same feature distance measure as Eq.~\ref{eq:lf}. In order to accelerate the feature simulation effect, we encourage minimizing the feature distance between adversarial and target images, meanwhile maximizing the feature distance from original image:
\begin{equation}
    r^* = \mathop{\arg\min}\limits_{r} \mathcal{L}_F(\hat{I}, I+r) - \mathcal{L}_F(I, I+r).
\end{equation}

\section{Experiment}
\label{sec:exp}
\begin{table*}[t!]
  \centering
  \resizebox{\textwidth}{!}{%
  \begin{tabular}{@{}l|c|c|c|c|c|c|c|c|c|c|c|c@{}}
    \toprule
    \diagbox{Attacks}{Models}& \multicolumn{3}{|c|}{SAM-B(white-box)} & \multicolumn{3}{|c|}{SAM-L} & \multicolumn{3}{|c|}{SAM-H} & \multicolumn{3}{|c}{FastSAM} \\
    \midrule
     Evaluation metric & mIoU$\downarrow$ & ASR@50$\uparrow$ & ASR@10$\uparrow$ & mIoU$\downarrow$ & ASR@50$\uparrow$ & ASR@10$\uparrow$ & mIoU$\downarrow$ & ASR@50$\uparrow$ & ASR@10$\uparrow$ & mIoU$\downarrow$ & ASR@50$\uparrow$ & ASR@10$\uparrow$ \\
    \midrule
    Attack-SAM-K \cite{zhang2023attacksam} &  $68.07\pm28.65$ & $24.10$& $6.87$ & $77.14\pm25.05$ & $14.46$ & $4.13$ & $78.71\pm24.02$ & $12.93$ & $3.51$ & $38.13\pm40.66$ & $59.90$ & $48.43$\\
    TAP \cite{zhou2018TAP} &  $63.49\pm32.58$ & $29.69$ & $13.12$ & $75.12\pm27.83$ & $16.96$ & $6.68$ & $77.36\pm26.21$ & $14.55$ & $5.31$ & $37.67\pm40.89$ & $60.53$ & $49.45$\\
    ILPD \cite{li2023ILPD} & $63.21\pm32.54$ & $30.15$ & $13.09$ & $75.17\pm27.75$ & $16.82$ & $6.59$ & $77.52\pm26.02$ & $14.37$ & $5.18$ & $37.84\pm40.84$ & $60.30$ & $49.02$\\
    AA \cite{Inkawhich2019AA} & $61.06\pm32.33$ & $32.48$ & $13.11$ & $70.70\pm29.74$ & $21.49$ & $8.67$ & $72.87\pm28.61$ & $19.13$ & $7.39$ & $32.64\pm39.58$ & $65.86$ & $55.10$\\
    PATA \cite{zheng2023pata} & $61.36\pm32.31$ & $32.23$ & $13.04$ & $70.81\pm29.73$ & $21.27$ & $8.62$ & $73.07\pm28.49$ & $18.66$ & $7.30$ & $32.74\pm39.56$ & $65.66$ & $54.97$\\
    PATA++ \cite{zheng2023pata} & $61.54\pm32.22$ & $32.00$ & $12.94$ & $71.02\pm29.55$ & $21.16$ & $8.38$ & $73.16\pm28.44$ & $18.84$ & $7.22$ & $32.85\pm39.60$ & $65.69$ & $54.65$\\
    \midrule
    UAD (ours) & $\mathbf{51.53\pm34.00}$ & $\mathbf{43.89}$ & $\mathbf{20.79}$ & $\mathbf{66.07\pm32.04}$ & $\mathbf{26.44}$ & $\mathbf{12.27}$ & $\mathbf{68.96\pm30.87}$ & $\mathbf{23.42}$ & $\mathbf{10.23}$ & $\mathbf{28.83\pm38.36}$ & $\mathbf{69.95}$ & $\mathbf{59.63}$\\
    \bottomrule
  \end{tabular}%
  }
  \caption{Results of our methods in comparison with prior and contemporary works.  Our proposed significantly outperforms other methods in both terms of average mask destruction (low mIoU) and number of drastically affected masks (high Attack Success Rate).}
  \label{tab:mainexp}
\vspace{-1em}
\end{table*}

\subsection{Experiment Settings}
\subsubsection*{Evaluation metrics}
We use three metrics to describe the effects of adversarial attack, which are mean Intersection over Union(mIoU), attack success rate at IoU$<50\%$ (ASR@50) and attack success rate at IoU$<10\%$(ASR@10). The first metric captures the average attack performance, and latter two capture how many output masks are significantly  destroyed by adversarial noise, which serve as worst case measurements. 

For robust evaluation to test the cross-prompt generalization, for each adversarial samples, we take randomness over prompts and report their mean and standard deviation of IoUs with annotated masks. For evaluating point prompts, we randomly sample 5 times for each grount-truth mask; for box prompts, we vary bounding box sizes for 3 times, resizing them to 80\% or 120\% of their original size.  

\vspace{.5em}
\noindent
\textbf{Compared baselines}

We carried out the experiments of our proposed attack in comparison with several prior or contemporary works: 
\begin{enumerate}
    \item Attack-SAM-K \cite{zhang2023attacksam} lowers the feature response globally given K (usually large, e.g. 400) point prompts over the whole image;
    \item \textit{Transferable Adversarial Perturbations} (TAP) \cite{zhou2018TAP} drives adversarial features away  from original features in Minkowski distance;
    \item \textit{Intermediate-level perturbation decay} (ILPD) \cite{li2023ILPD} is a refined version of TAP, keeping an effective adversarial direction while possessing a greater magnitude;
    \item \textit{Activation attack} (AA) \cite{Inkawhich2019AA} minimizes the distance between the adversarial feature and a target image feature;
    \item  \textit{Prompt-Agnostic target attack} (PATA) \cite{zheng2023pata} introduces a regularization term to boost the feature dominance of adversarial image over a random clean competition image, on the basis of AA \cite{Inkawhich2019AA}.
    \item PATA++ \cite{zheng2023pata} is an enhanced version of PATA that addresses the conflict between increasing feature similarity and reducing feature dominance. PATA++ alleviates the issue by randomly pick one new competition image in every adversarial update iteration.  
\end{enumerate}

\vspace{.5em}
\noindent \textbf{Attack settings}

In all the experiments, we set adversarial update steps of final adversarial example to be 40. The results shown in Fig. \ref{fig:examplars} are adversarial examples crafted using a mild  $\epsilon = 12/255$ noise to highlight the attack results. In other experiments, if not explicitly stated, the perturbation range $\epsilon$is set to $ 8/255$ and perturbation step size $\alpha$ is set to $2/255$. For our attack, we set the proxy perturbation iterations$T_f$ to be $4$, allowing the proxy example to approximately reach the feasible set boundary. The deformation iteration is set to $40$. 

\subsection{Adversarial Examples}
We show the effectiveness of our method through some adversarial examples in Fig.\ref{fig:examplars}. We have chosen art paintings\footnote{\href{https://www.kaggle.com/code/paultimothymooney/collections-of-paintings-from-50-artists/notebook}{Kaggle Artist dataset}}, sculpture designs\footnote{\href{https://www.kaggle.com/datasets/thedownhill/art-images-drawings-painting-sculpture-engraving}{Art Images dataset}}, and personal photos\footnote{\href{https://www.kaggle.com/datasets/hsankesara/flickr-image-dataset}{FLICKR30K dataset}} to underscore the practical usage of the \textit{Anything Unsegmentable} task to safeguard digital assets, art copyrights, and portrait rights.  All the advresarial samples are crafted on SAM-B model, but they indeed transfer to SAM-H and FastSAM models, regardless of their changes in parameters or architectures.

Upon examining the results presented in Figure \ref{fig:examplars}, it is evident that our attack has a profound impact on the destruction of masks, regardless of whether the prompts are spatial or semantic. In white-box scenarios where the target model is SAM-B, the alterations are most noticeable, with the bounding box prompt highlighting little more than meaningless ripples on the image. Even for black-box models SAM-H and FastSAM, the segmentation masks are significantly distorted. The point prompt and text prompt, which originally could highlight the entire foreground object, now, due to the influence of the deformation target, highlights disjointed parts that cannot form a valid whole object. Importantly, we can observe clear evidence that segmentation results are influenced by the deformed target. For example, in the rightmost figure of the second row, the text prompt "Marble" highlights an area that was originally a background in the clean image. Notably, this wrong segmentation aligns with the deformed target image, where a part of a marble statue is displayed.

\subsection{Quantitative Evaluation}
We conducted a comprehensive comparison of our approach with all prior works in Tab. \ref{tab:mainexp}. All adversarial examples were generated from SAM-B, which has the smallest parameter size among the SAM family of models. Consequently, the other three models (SAM-L, SAM-H, and FastSAM) are considered as black-box models. We selected the SAM-1B dataset for evaluation \cite{kirillov2023segment}, which includes a wide range of diverse images close to real-world scenarios. We conducted our study on a  subset of the SAM-1B, specifically the first 1000 images (sa\_1.jpg to sa\_1000.jpg in subset sa\_000000.tar). This subset encompasses a total of 98875 masks, which is already a large quantity of masks and has statistical significance.

The results in Table \ref{tab:mainexp} reveal several intriguing facts. Firstly, our proposed attack achieves state-of-the-art results and outperforms other methods by a large margin. Secondly, upon comparing different methods, we observe that targeted feature disruption attacks (AA \cite{Inkawhich2019AA} and PATA \cite{zheng2023pata}) perform significantly better than untargeted feature attacks (TAP \cite{zhou2018TAP} and ILPD \cite{li2023ILPD}). This aligns with our earlier analysis, which indicated that optimizing within the feature manifold yields better results than moving away from it. Attack-SAM-K \cite{zhang2023attacksam} exhibits the weakest performance, further supporting the notion that prompt-specific attacks are less effective than feature perturbation attacks. 

Interestingly, we observed that FastSAM performs notably poor, exhibiting a very low mean Intersection over Union (mIoU) and a high portion of masks with drastic changes. This subpar performance might be attributed to the fact that the authors of FastSAM \cite{zhao2023fastsam} trained their model using only 2\% of the SAM-1B dataset. The limited training data likely led to a lack of robustness in their model
, making it sensitive to out-of-distribution samples.

\subsection{Ablation Studies}

\subsubsection{On Perturbation Budget $\epsilon$}
We conducted experiments using different values of $\epsilon$ to assess the effectiveness of our proposed attack across varying perturbation ranges. Remarkably, even when operating within a small perturbation range ($ \epsilon = 4 $), UAD significantly outperforms other perturbation methods. This demonstrates the superiority and versatility of our approach across a wide range of scenarios.

\begin{figure}[t!]
  \centering
   \setlength{\belowcaptionskip}{-1em}
   \includegraphics[width=\linewidth]{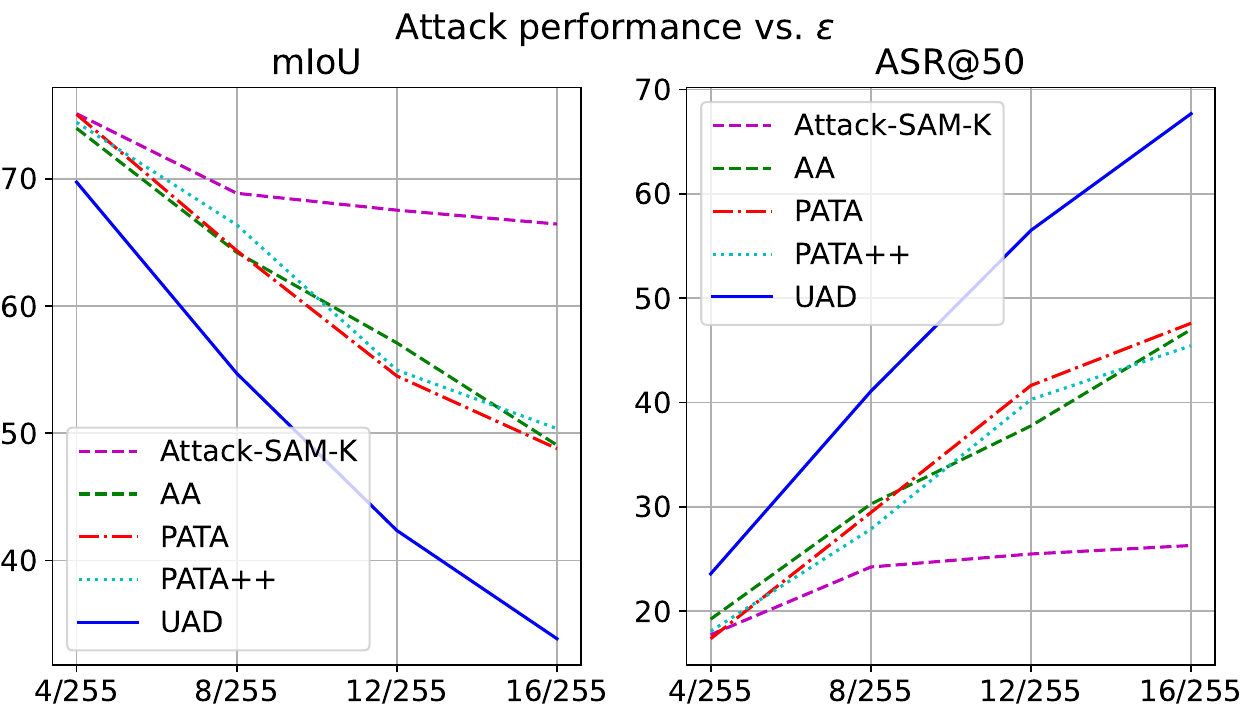}
   \caption{Our attack show consistent superiority under different settings of attack perturbation range $\epsilon$.}
   \label{fig:epsilon}
\end{figure}
\vspace{-0.5em}

\subsubsection{On Functionality of Loss Terms}
We show all loss components in our optimization process $ \mathcal{L} = \mathcal{L}_D + \lambda_C \mathcal{L}_C + \lambda_F \mathcal{L}_F$ are necessary and complementary. High-level speaking, first two terms controls deformation to be structural dissimilar($\mathcal{L}_D$) and locally smooth($\mathcal{L}_C$), while last term $\mathcal{L}_F$ constrains adversarial feature distance.

\begin{figure}[ht]
    \centering
    \begin{subfigure}[b]{0.1\textwidth}
        \includegraphics[width=\textwidth]{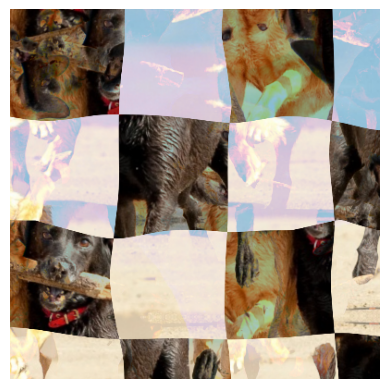}
        \caption{UAD}
        \label{fig:image1}
    \end{subfigure}
    \hfill
    \begin{subfigure}[b]{0.1\textwidth}
        \includegraphics[width=\textwidth]{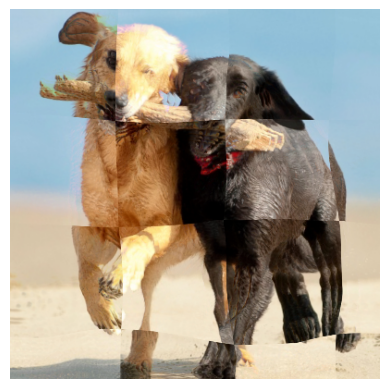}
        \caption{no $\mathcal{L}_D$}
        \label{fig:image2}
    \end{subfigure}
    \hfill
    \begin{subfigure}[b]{0.1\textwidth}
        \includegraphics[width=\textwidth]{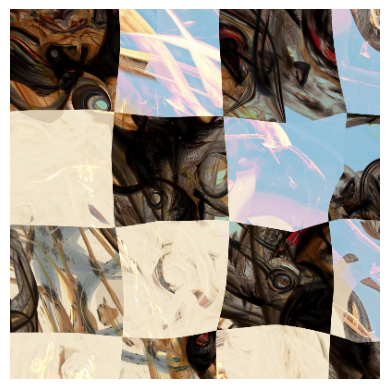}
        \caption{no $\mathcal{L}_C$}
        \label{fig:image3}
    \end{subfigure}
    \hfill
    \begin{subfigure}[b]{0.1\textwidth}
        \includegraphics[width=\textwidth]{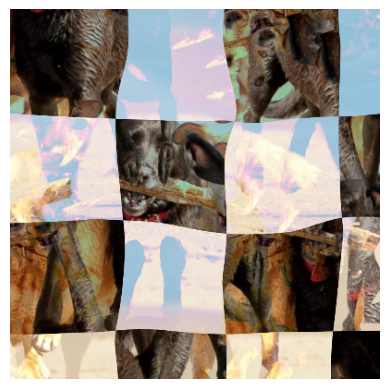}
        \caption{no $\mathcal{L}_F$}
        \label{fig:image4}
    \end{subfigure}
    \label{fig:images}
\caption{Deformed targets without each individual loss terms.}
\label{fig:ablossterm}
\end{figure}
\vspace{-1em}

We visualize deformed targets in Fig.~\ref{fig:ablossterm} for cases without individual loss terms and illustrate the functionality of each loss term vividly. Without $\mathcal{L}_D$, there would be almost no structural change to misguide segmentation models. Without $\mathcal{L}_C$, the deformation inside each patch goes too wild to contain valid natural shape. This will cause a target image away from the image manifold, resulting in sub-optimal attacks. Without $\mathcal{L}_F$ the attack is less effective, which we will further illustrate in the next subsection.

\vspace{-1em}
\subsubsection{On Proxy Adversarial Update Steps $T_f$}
In each deformation step, we calculate the feature distance between the deformed target and a proxy adversarial sample as an estimation of the feature distance from deformed target to the feasible set of adversarial images. 
We have observed that increasing the number of iterations leads to a more reachable target from the feasible set, allowing for better control over the deformation's development. However, it's essential to strike a balance between achieving finer adversarial attack performance and incurring a higher computational overhead. 

\begin{figure}[t!]
  \centering
   \setlength{\abovecaptionskip}{-0.05cm}
   \includegraphics[width=0.9\linewidth]{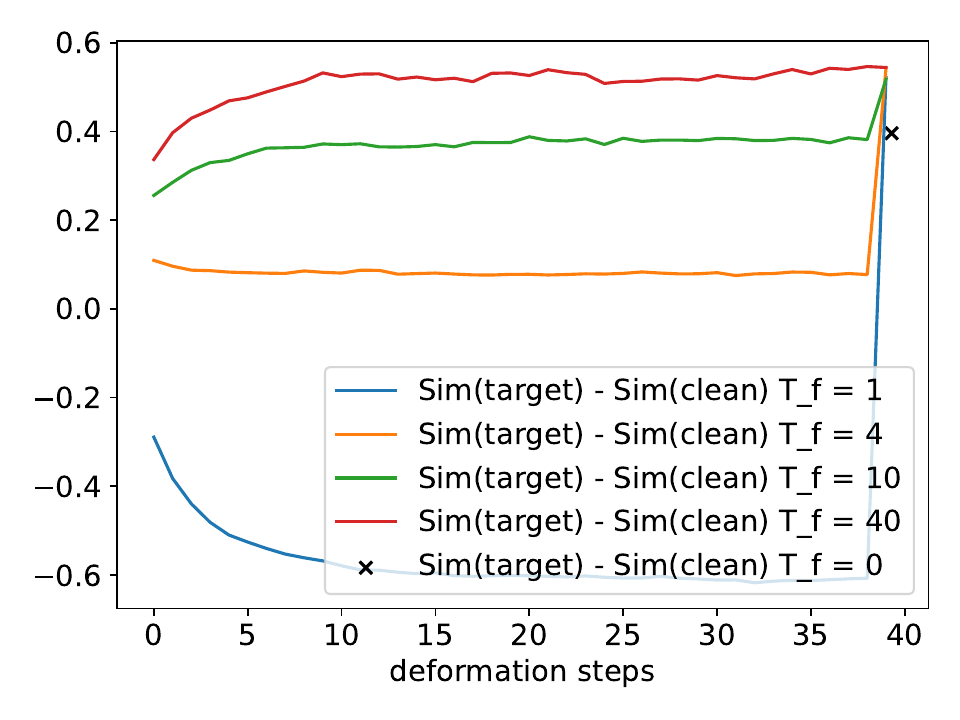}
   \caption{The development of relative feature similarity during deformation steps. More proxy adversarial iterations result in a closer feature distance between deformed target and adversarial example, coming at the cost of heavier computation burden.}
   \label{fig:fidelity}
   \vspace{-0.5em}
\end{figure}

The ablation study on $T_f$ presented in Fig.~\ref{fig:fidelity} underscore the importance of incorporating fidelity loss. We illustrate the changes in relative feature similarity, which is defined as the feature similarity of $I+r$ to $\hat{I}$ minus the feature similarity between $I+r$ and $I$. Our primary focus here lies in the last column of the plotted figure, when the deformation target has been optimized and fixed, and the final adversarial example is obtained by $T$ steps to simulate the deformation. The variations in values within the last column indicate how closely the adversarial image can approach the target by the end of the optimization process.

When not using any proxy adversarial samples when updating deformation, the relative feature similarity is obviously lower (black mark in the plot). With more proxy adversarial updates integrated into the process, the final relative feature similarity increases. We found that increasing proxy adversarial iterations has diminishing returns. Compared to using the actual attack step count ($T_f = T = 40$) to estimate the distance at each step, using fewer steps only marginally reduces relative feature similarity. 
In practice, to avoid introducing excessive computational burden, we choose $T_f = 4$ as a balanced point that maintains simulation effectiveness while ensuring efficiency.

We put more ablation studies in the appendix.

\section{Related Work}
\label{sec:related_work}
\noindent\textbf{Promptable Segmentation Models.}
In recent developments of segmentation techniques, there has been a shift from closed-set, non-interactive segmentations~\cite{he2017mask, chen2017deeplab, kirillov2019panoptic,SuchengCVPR22,SuchengICCV23,MetaformerBaseline} to more open-vocabulary and interactive settings. This evolution allows for a wide range of interaction forms, including clicks~\cite{Xu_Price_Cohen_Yang_Huang_2016, chen2022focalclick, liu2023simpleclick}, bounding boxes~\cite{kirillov2023segment}, scribbles~\cite{zou2023segment}, text~\cite{ghiasi2022scaling, ding2022open, xu2022groupvit, xu2023open, kirillov2023segment}, or contextual information~\cite{wang2023seggpt, zou2023segment}. Notably, Segment Anything Model (SAM)~\cite{kirillov2023segment} has the most remarkable zero-shot ability due to its massive training dataset containing 11 million images and 1.1 billion masks. Scaling up training data consequently result in a highly generalized and robust model~\cite{kirillov2023segment, samrobust1, samrobust2}, that nowadays people call them \textit{Vision Foundation Models}. Subsequent research improved SAM in terms of quality \cite{ke2023samhq}, latency
\cite{zhao2023fastsam} and semantic awareness \cite{li2023semantic}. 

\noindent\textbf{Adversarial Attacks for Segmentation Models.}
While most adversarial attack research focuses on classifiers, some work extends these attacks to segmentation models. \citep{arnab2018robustness} evaluated adversarial attacks on semantic segmentation models, finding that these models are more robust than classifiers due to their multi-scale processing. \cite{xie2017adversarial} introduced Dense Adversary Generation, encouraging incorrect recognition on multiple targets simultaneously. \cite{hendrik2017universal} proposed generating universal perturbations guiding networks to create desired target segmentations. \cite{chen2022semantically} discussed stealthy attacks on segmentation models, altering targeted labels while keeping non-targeted labels intact. \cite{gu2022segpgd} suggested an efficient attack reducing the number of iterations. These methods mainly focus on closed-vocabulary, non-promptable segmentation models for per-pixel classification. Very recently, we notice there are also some work discussing the adversarial attacks on SAM \cite{zhang2023attacksam, samrobust1, samrobust2, han2023samuap, zheng2023pata}. \citet{zhang2023attacksam} designed the first adversarial attack towards SAM which optimizes the input image to have negative feature responses in mask area. SAM-UAP \cite{han2023samuap} introduces universal adversarial perturbations towards SAM from a contrastive learning perspective. \cite{zheng2023pata} propose Prompt-Agnostic Targetted Adversarial Attacks(PATA) to generate more transferable samples. In our paper, we selected Attack-SAM and PATA as two important baselines and show that our method exhibits higher effectiveness and transferability.

\noindent\textbf{Transferability of Adversarial Attacks.}
Adversarial examples often struggle to transfer successfully across various neural networks, yielding low success rates in black-box settings. Previous research has explored methods to craft more transferable adversarial samples. These approaches encompass techniques like applying gradient momentum \cite{dong2018momentum, Lin2020Nesterov, wang2021boosting}, input augmentation \cite{xie2019improving, dong2019evading, wang2021enhancing, long2022frequency, wang2021admix}, feature disturbance \cite{ganeshan2019fda, huang2019enhancing, zhou2018TAP, wang2021feature, wu2020adatt, Inkawhich2019AA, li2023ILPD}, and model ensembling \cite{Li2020ghost, liu2016delving}. For a comprehensive and up-to-date survey, readers are referred to \cite{zhao2023revisiting}. These research directions offer various strategies for boosting the transferability of adversarial attacks~\cite{JingwenAAAI24}. 

\section{Conclusion}
In this study, we present a novel challenge termed \textit{Anything Unsegmentable}, aimed at generating highly transferable, prompt-agnostic adversarial examples. These examples are designed to shield individuals from the potential risks of copyright and privacy violations posed by foundational promptable segmentation models. Our method \textit{Unsegment Anything by Simulating Deformation} (UAD), marks a progressive step towards addressing this challenge, outperforming existing and concurrent approaches. Our analysis of the robustness of foundational segmentation models uncovers two compelling insights: (1) prompt-specific attacks struggle with transferability, and (2) targeted feature perturbations towards natural-image-like samples, are significantly more effective than untargeted perturbations that drive features away from their original location. We hope that our work will provide valuable perspectives on the resilience of these powerful vision models and inspire future research to mitigate the societal issues they may engender.

\section{Acknowledgement}
This project is supported by the National Research Foundation, Singapore under its AI Singapore Programme (AISG Award No: AISG2-RP-2021-023), and the Singapore Ministry of Education Academic Research Fund Tier 1 (WBS: A-0009440-01-00).


\clearpage
\setcounter{page}{1}
\maketitlesupplementary
\setcounter{section}{0}

\newcommand{\yxy}[1]{\textcolor{orange}{\emph{[yxy: #1]}}}
\newcommand{\yxyc}[1]{\textcolor{gray}{\emph{[yxyc: #1]}}}

The supplementary material is organized as follows:
In Sec. \ref{sec:prompt-specific}, we present our findings regarding the challenges of transferring prompt-specific attacks to unseen prompts. We offer additional ablation studies in Sec. \ref{sec:ablation-appendix}, covering topics including (1) the combination with other transferability methods, and (2) the impact of source models. These sections aim to enhance the reader's comprehension of our approach's underlying mechanisms. In Sec. \ref{sec:vis}, we offer visualizations of our attacks and baseline attacks from a panoramic perspective to provide a more straightforward comparison. In Sec. \ref{sec:notation}, we provide a table listing all the notations used throughout this paper. Finally, in Sec. \ref{sec:conclusion}, we discuss the limitations of our work and the potential societal impact that may arise from our research.

\section{Prompt-Specific Attack Fail to Transfer}
\label{sec:prompt-specific}

\begin{figure*}[ht!]
  \centering
   \includegraphics[width=\linewidth]{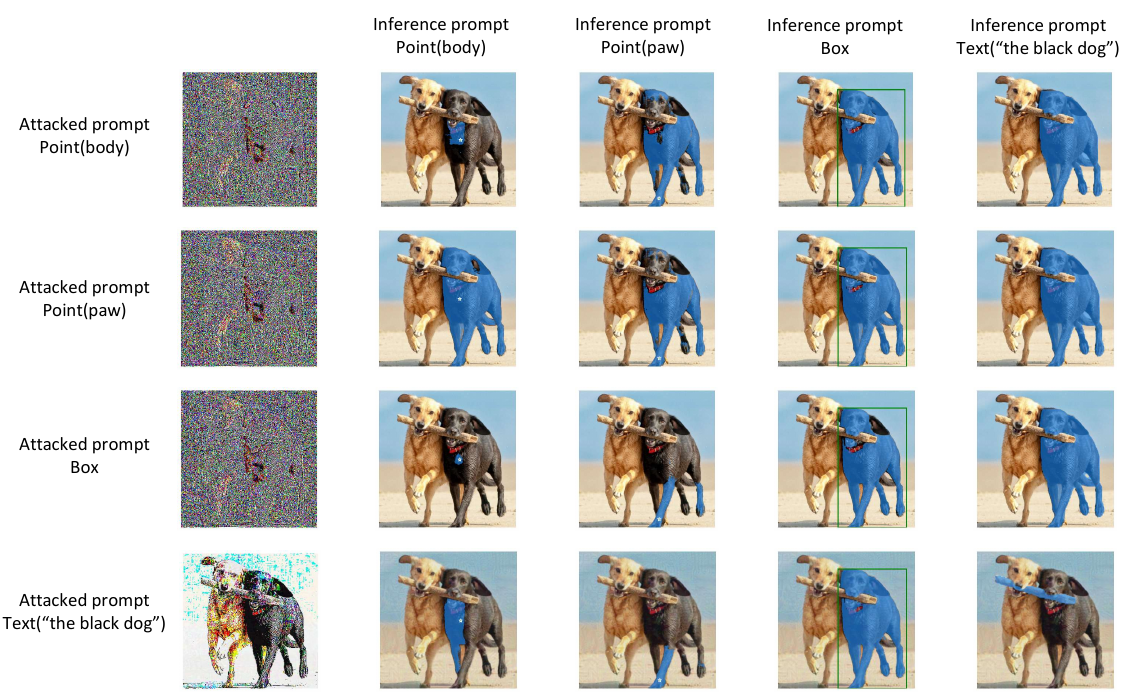}
   \caption{We claim that prompt-specific attacks exhibit fundamental differences in the adversarial noise they generate, and their transferability is limited to a narrow range of prompts. Adversarial examples tend to overfit to the prompts used during the attack phase and have limited impact on unseen prompts.}
   \label{fig:untransfer}
\end{figure*}

We present visualizations of adversarial noises alongside their corresponding segmentation results for various prompts in Fig. \ref{fig:untransfer}. These visualizations underscore the heterogeneous nature of prompt-specific adversarial noises.

Notably, the adversarial noises induced by spatial prompts (the first three rows) exhibit distinct characteristics compared to those induced by semantic prompts (the last row). The adversarial noises from the spatial prompts share similarities, characterized by shattered and random-noise-like patterns. Conversely, the adversarial noise stemming from the semantic prompt aligns closely with essential image features.

Furthermore, we observe that the generated adversarial examples tend to exhibit overfitting to the specific prompts used during their generation. Consequently, these adversarial examples struggle to generalize to unseen prompts. Attacks generated using spatial prompts have minimal impact on segmentation results driven by text prompts. Similarly, the adversarial sample generated from a text prompt has little effect on box prompts.

\noindent\textbf{Results.} We present the histogram of feature similarities between TAP and AA attacks in Fig.~\ref{fig:untarget}.
The findings reveal that both targeted and untargeted feature disruption attacks effectively alter features in the source model. However, untargeted attacks are notably less effective on the target model. We hypothesize this is due to the high-dimensional nature of the data, which often leads these attacks to stray from the image distribution. As a result, in the target model, adversarial features appear similar to normal features, indicating lower transferability for untargeted attacks.


\section{Ablation Studies}
\label{sec:ablation-appendix}

\begin{table*}[h!]
  \centering
  \resizebox{\textwidth}{!}{%
  \begin{tabular}{@{}l|c|c|c|c|c|c|c|c|c|c|c|c@{}}
    \toprule
    \diagbox{Approach}{Models}& \multicolumn{3}{|c|}{SAM-B (source)} & \multicolumn{3}{|c|}{SAM-L} & \multicolumn{3}{|c|}{SAM-H} & \multicolumn{3}{|c}{FastSAM} \\
    \midrule
     & mIoU$\downarrow$ & ASR@50$\uparrow$ & ASR@10$\uparrow$ & mIoU$\downarrow$ & ASR@50$\uparrow$ & ASR@10$\uparrow$ & mIoU$\downarrow$ & ASR@50$\uparrow$ & ASR@10$\uparrow$ & mIoU$\downarrow$ & ASR@50$\uparrow$ & ASR@10$\uparrow$ \\
    \midrule
    UAD &  \textbf{51.53} & 43.89 & 20.79 & \textbf{66.07} & \textbf{26.44} & \textbf{12.27} & \textbf{68.96} & \textbf{23.42} & \textbf{10.23} & 28.83 & 69.95 & 59.63\\
    UAD + DI \cite{xie2019improving} & 56.74 & 41.01 & 18.51 & 70.19 & 24.54 & 11.07 & 71.25 & 22.15 & 8.83 & 28.88 & 65.38 & 56.69 \\
    UAD + MI \cite{dong2018momentum} & 55.99 & \textbf{44.29} & \textbf{21.54} & 67.79 & 25.29 & 11.80 & 69.62 & 23.22 & 9.97 & \textbf{27.44} & \textbf{70.97} & \textbf{59.81} \\
    UAD + MI \cite{dong2018momentum} + DI \cite{xie2019improving} & 56.49 & 42.06 & 19.24 & 67.94 & 24.76 & 11.32 & 69.26 & 23.03 & 9.51 & 27.60 & 70.36 & 59.58\\
    \bottomrule
  \end{tabular}%
  }
  \caption{Results of combining our method with gradient momentum and input augmentation.}
  \label{tab:DIMI}
\end{table*}

\subsection{Study 1: Combining Transferability Methods}
Previous research on the transferability of adversarial examples has highlighted four distinct technical approaches, as discussed in Section \ref{sec:related_work}. These approaches encompass feature disturbance, gradient momentum, input augmentation, and model ensembling. Notably, our UAD method falls under the category of feature disturbance.  In the precondition of not  introducing external model information (we will investigate the impact of model ensembling in the next subsection), we concentrate on exploring how gradient momentum and input augmentation could potentially help us to reach our objective.

As recently demonstrated in a comprehensive benchmark study \cite{zhao2023revisiting}, under a fair and rigorous comparison, the most effective gradient momentum and input augmentation methods are, in fact, the most classic ones, specifically MI \cite{dong2018momentum} and DI \cite{xie2019improving}, respectively. In Table \ref{tab:mainexp}, we have already presented results indicating that the inclusion of both of these techniques does not significantly enhance attack performance. Now we will separately integrate each of these techniques into our method, given that they should operate independently of each other, and examine their combined effects in conjunction with our proposed approach in Tab. \ref{tab:DIMI}.

Contrary to our expectations, our method, when used alone without the inclusion of MI or DI tricks, yielded the best results. The addition of gradient momentum proved to be more effective than data augmentation. However, combining both techniques resulted in a drop in performance. 

\subsection{Study 2: Source Model Selection}
Previous research findings provide valuable insights: (1) Adversarial examples generated by high-capacity (more over-parameterized) models exhibit higher transferability to low-capacity networks, in contrast to adversarial samples crafted by low-capacity networks, which have limited success when transferred to high-capacity networks; (2) Employing an ensemble of networks proves to be more effective in generating transferable adversarial samples.

To evaluate the upper limits of our approach in tackling the \textit{anything unsegmentable} task, we conducted an ablation study to assess the enhanced transferability of adversarial examples generated from a more capable model.

In Table \ref{tab:mainexp}, we performed all experiments using SAM-B as the source model, which has one of the smallest parameter sizes (91 M). Now, we aim to generate adversarial samples based on larger and more powerful models, such as SAM-L (308 M parameters) and SAM-H (636 M parameters). Additionally, we conducted experiments by ensembling SAM-B and SAM-L. We couldn't ensemble SAM-H due to its high GPU memory consumption, as we have limited computational resources.

\begin{table*}[t!]
  \centering
  \resizebox{\textwidth}{!}{%
  \begin{tabular}{@{}l|c|c|c|c|c|c|c|c|c|c|c|c@{}}
    \toprule
    \diagbox{Source Models}{Target Models}& \multicolumn{3}{|c|}{SAM-B} & \multicolumn{3}{|c|}{SAM-L} & \multicolumn{3}{|c|}{SAM-H} & \multicolumn{3}{|c}{FastSAM} \\
    \midrule
     & mIoU$\downarrow$ & ASR@50$\uparrow$ & ASR@10$\uparrow$ & mIoU$\downarrow$ & ASR@50$\uparrow$ & ASR@10$\uparrow$ & mIoU$\downarrow$ & ASR@50$\uparrow$ & ASR@10$\uparrow$ & mIoU$\downarrow$ & ASR@50$\uparrow$ & ASR@10$\uparrow$ \\
    \midrule
    SAM-B (91 M) &  51.53 & 43.89 & 20.79 & 66.07 & 26.44 & 12.27 & 68.96 & 23.42 & 10.23 & 28.83 & 69.95 & 59.63\\
    
    SAM-L (308 M) & 61.67 & 35.65 & 13.45 & \textbf{55.41} & \textbf{28.50}  & \textbf{15.53} & 68.98  & 23.37 & 10.41 & 31.27 & 63.23 & 52.27 \\
    
    SAM-H (636 M) & 61.06 & 35.69 & 13.87 & 63.92 & 24.32 & 13.04 & \textbf{63.31} & \textbf{25.85} & \textbf{12.61} & 30.60 & 64.59 & 53.75 \\
    SAM-B + SAM-L (Ensemble) & \textbf{50.54} & \textbf{44.18} & \textbf{22.61} & 59.67 & 26.88 & 14.40 & 68.35 &  23.81 & 11.35 & \textbf{27.36} & \textbf{71.38} & \textbf{60.90} \\
    \bottomrule
  \end{tabular}%
  }
  \caption{Ablation study on source models used to craft adversarial examples. }
  \label{tab:atksrc}
\end{table*}

As indicated in Table \ref{tab:atksrc}, our attack, much like many other adversarial attacks, demonstrates a tendency to overfit to the source model. Specifically, when the source and target models are identical, the attack performs significantly better when the source model does not encompass the target model. Interestingly, model ensembling further enhances attack results; for instance, ensembling SAM-B and SAM-L surpasses the performance of using SAM-B alone by a considerable margin. Ensembling exhibits a stronger impact on the global results, leading to a lower mean IoU (mIoU). 

\section{Algorithm Pseudo-code}
We present the pseudo-code of our attack in Alg.\ref{alg:UAD}. 

\begin{algorithm}[h!]
    \centering
    \caption{Unsegment Anything by Simulating Deformation}
    \label{alg:UAD}
    \begin{algorithmic}[1]
        \Require \begin{tabular}[t]{@{}l}
            Input image: $I$; \\
            Deformation parameters: $w$; \\
            Maximal deformation iterations: $T_D$; \\
            Maximal proxy perturbation iteration: $T_f$; \\ 
            Maximal perturbation iterations: $T$; \\
            Perturbation step size: $\alpha$; \\
            Perturbation range: $\epsilon$;
          \end{tabular}
        \Ensure {Adversarial perturbation: $r$}  
        \Procedure{UAD}{$I, w, T_D, T_f, T, \alpha, \epsilon$}
        \State $I' = I$, $r = 0$, $t_D = 0$, $t = 0$;
        \State Initialize $w$ so that $D_w$ produces identity mapping;
        \State \textbf{While} $t_D < T_D$ \textbf{do}   \Comment{Stage 1: Deformation}
        \State \quad $\hat{I} = \mathcal{D}_w(I)$; \Comment{Get deformed image}
        \State \quad $I'' = I$; 
        \State \quad $t_f = 0$;
        \State \quad \textbf{While} $t_f < T_f$ \textbf{do} \Comment{proxy adversarial sample}
        \State \quad\quad $I'' = I'' - \alpha\cdot sign(\nabla_{I''}\mathcal{L}_F (\hat{I}, I''))$;
        \State \quad\quad $I'' = clip_{\epsilon}(I'' - I) + I$
        \State \quad\quad $I'' = clip_{0, 1}(I'') $
        \State \quad \textbf{end While}
        \State \quad $\mathcal{L}(w) = \mathcal{L}_D(\hat{I}, I) + \mathcal{L}_C (w) + \mathcal{L}_F(\hat{I}, I'')$
        \State \quad $w = w - \nabla_w \mathcal{L}(w) $  \Comment{Update deformation parameters}
        \State \textbf{end While}
        \State $\hat{I} = \mathcal{D}_w(I)$;  \Comment{End of stage 1, deformed target fixed}
        \State \textbf{While} $t < T$ \textbf{do}   \Comment{Stage 2: Simulation}
        \State \quad $I' = I' - \alpha \cdot sign(\nabla_{I'}\mathcal{L}_F (\hat{I}, I'))$
        \State \quad $I' = clip_{\epsilon}(I' - I) + I$
        \State \quad $I' = clip_{0, 1}(I')$
        \State \textbf{end While}
        \State $r = I' - I$
        \State \textbf{return} $r$
        \EndProcedure
    \end{algorithmic}
\end{algorithm}

\section{Notations}
\label{sec:notation}
We put all symbols appeared in this paper in Tab. \ref{tab:symbols} for reference.
\begin{table}[h]
\centering
\caption{Notation Table}
\label{tab:symbols}
\begin{tabular}{ll}
\toprule
\textbf{Variable} & \textbf{Description} \\
\midrule
$I$ & Original clean image \\
$P$ & Prompt \\
$M$ & Mask \\
$f_{\theta^I}$ & Image encoder of the promptable segmentation model \\
$h_{\theta^P}$ & Prompt encoder of the promptable segmentation model \\
$g_{\theta^M}$ & Prompt encoder of the promptable segmentation model \\
$w$ & Deformation control parameters\\
$\mathcal{D}_w$ & Deformation function \\
$\hat{I}$ & Deformed (target) image \\
$w_{ff}$ & Parameters of flow field which controls deformation \\
$w_{ff}^{(i, j)}$ & Flow vector of position $(i, j)$ in $w_{ff}$ \\
$\nabla u$ & Movement in width indicated by flow field\\
$\nabla v$ & Movement in height indicated by flow field\\
$r$ & Adversarial perturbation (adversarial noise) \\
$I' = I + r$ & Adversarial image \\
$T_D$ & Deformation iterations \\
$T_f$ & Proxy adversarial update iterations \\
$T$ & Adversarial update iterations \\
$\alpha$ & Adversarial perturbation step size \\
$\epsilon$ & Adversarial perturbation range \\
$\mathcal{L}_D$ & Deformation loss\\
$\mathcal{L}_C$ & Control loss\\
$\mathcal{L}_F$ & Fidelity loss \\
\bottomrule
\end{tabular}

\end{table}

\section{More visualization results}
\label{sec:vis}
\begin{figure*}[h!]
    \centering
    \begin{subfigure}[b]{0.88\textwidth}
        \includegraphics[width=\textwidth]{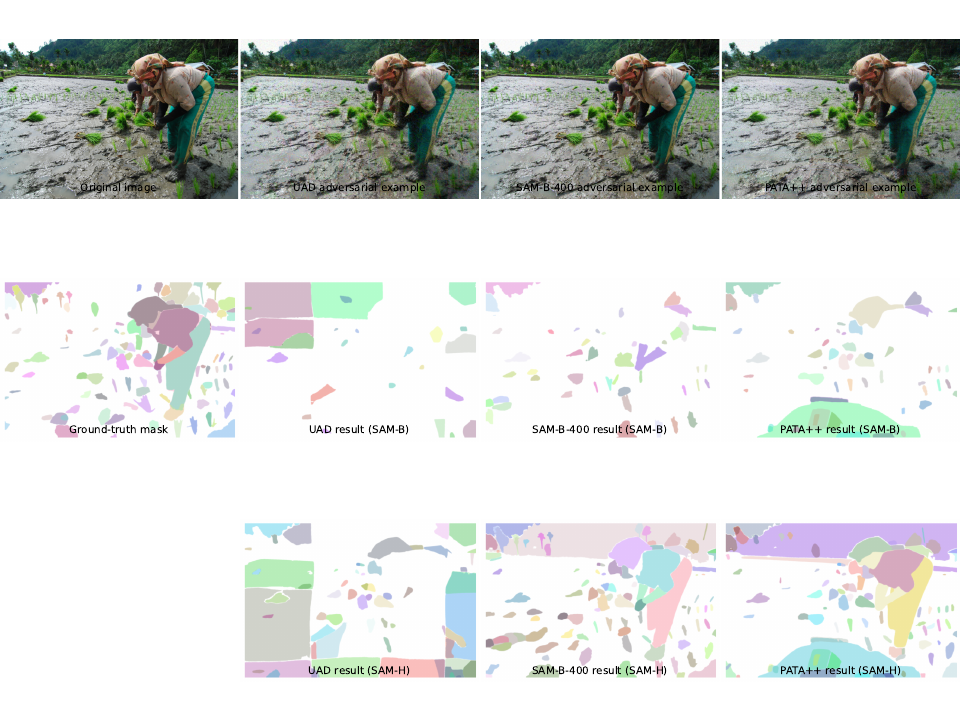}
        \label{fig:subfig1}
    \end{subfigure}
    \hfill
    \begin{subfigure}[b]{0.88\textwidth}
        \includegraphics[width=\textwidth]{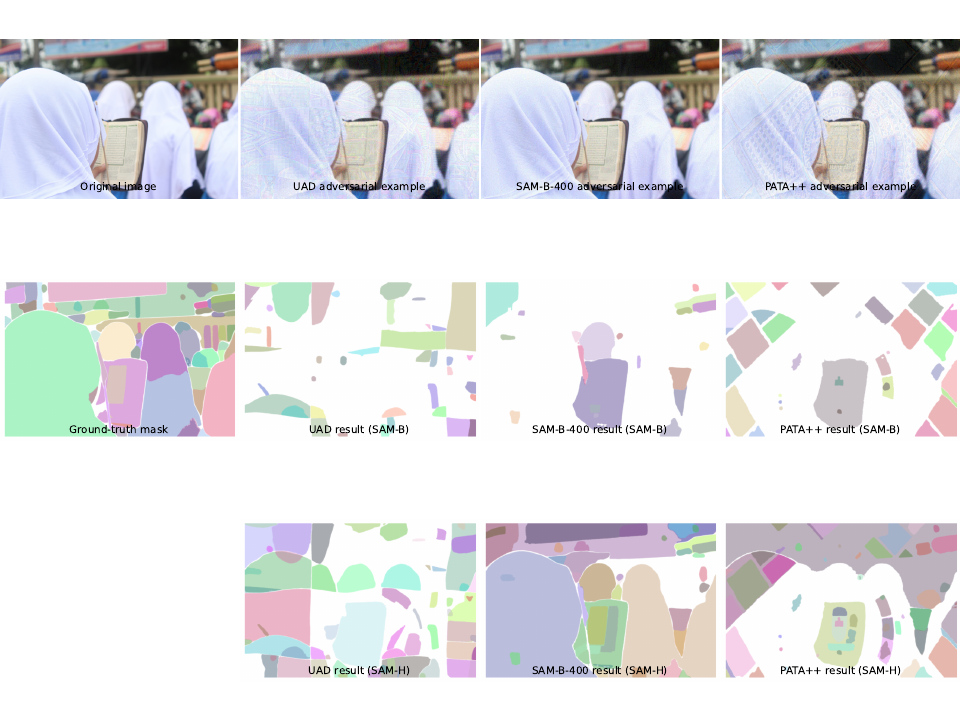}
        \label{fig:subfig2}
    \end{subfigure}
    \caption{Visualizations of attack results in panoramic view(I) }
    \label{fig:pano1}
\end{figure*}

\begin{figure*}[h!]
    \centering
    \begin{subfigure}[b]{0.88\textwidth}
        \includegraphics[width=\textwidth]{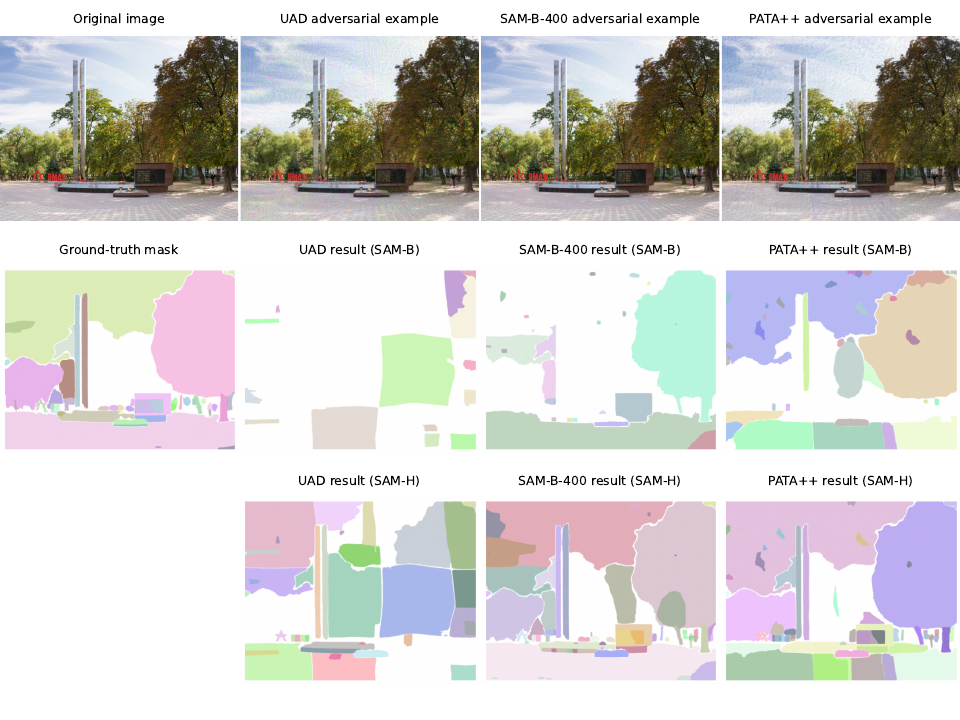}
        \label{fig:subfig1}
    \end{subfigure}
    \hfill
    \begin{subfigure}[b]{0.88\textwidth}
        \includegraphics[width=\textwidth]{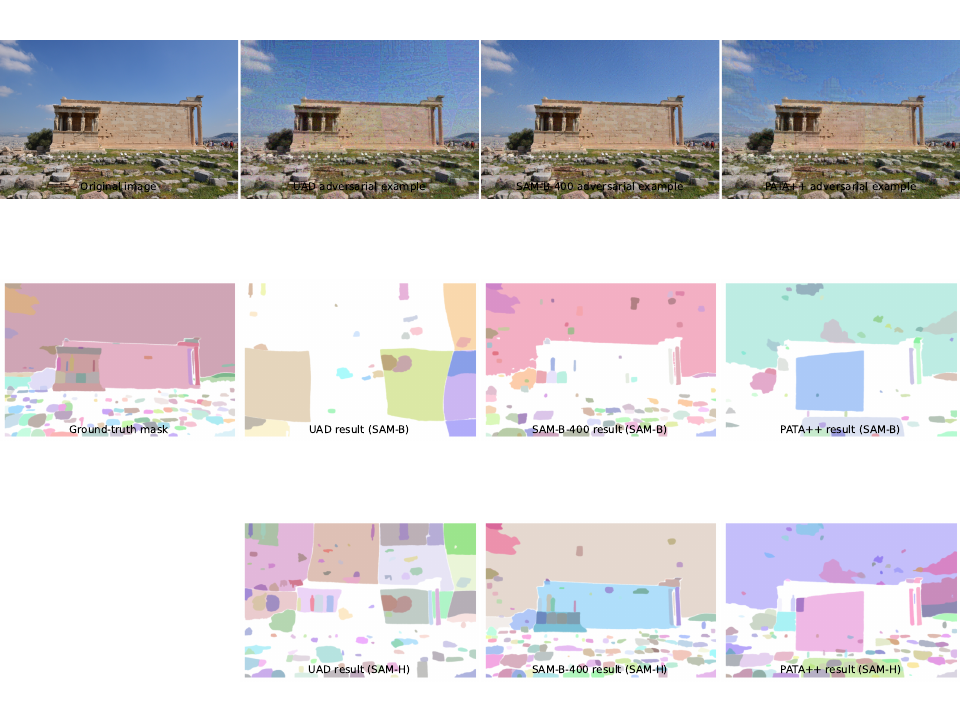}
        \label{fig:subfig2}
    \end{subfigure}
    \caption{Visualizations of attack results in panoramic view(II) }
    \label{fig:pano2}
\end{figure*}

\begin{figure*}[h!]
    \centering
    \begin{subfigure}[b]{0.9\textwidth}
        \includegraphics[width=\textwidth]{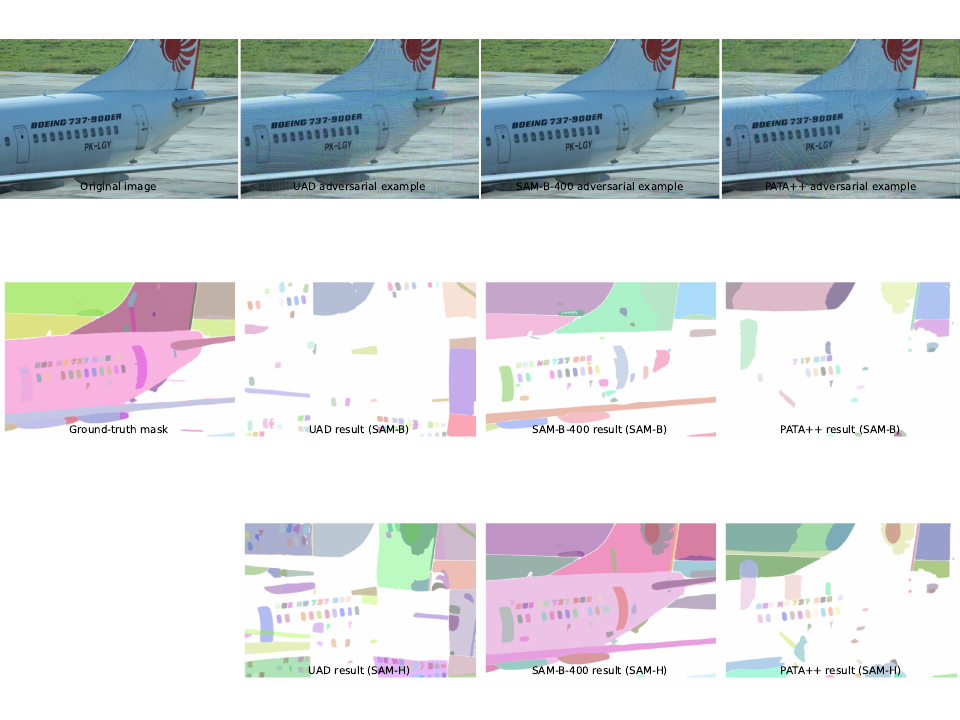}
        \label{fig:subfig1}
    \end{subfigure}

    \begin{subfigure}[b]{0.9\textwidth}
        \includegraphics[width=\textwidth]{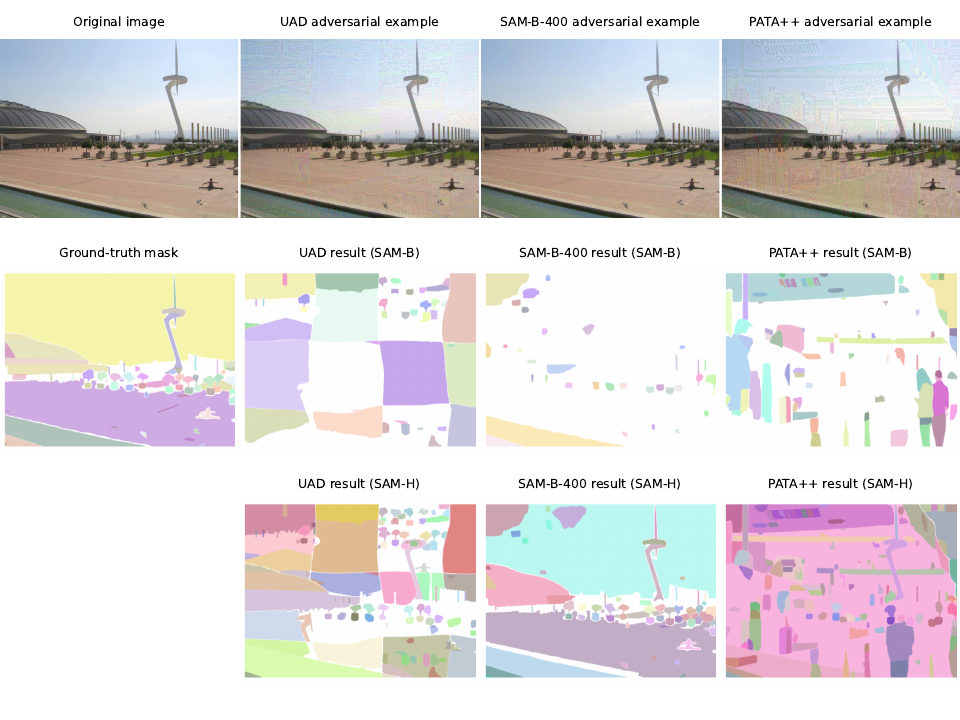}
        \label{fig:subfig2}
    \end{subfigure}
    \caption{Visualizations of attack results in panoramic view(III) }
    \label{fig:pano3}
\end{figure*}

We visualize the attack effect under segment everything mode (which provides a panoramic view without prompts) on SAM-B and SAM-H models in Fig.\ref{fig:pano1}, Fig. \ref{fig:pano2} and \ref{fig:pano3}. We compare the attack results with Attack-SAM and PATA++ to highlight the difference in failure patterns and our effectiveness.

\clearpage
\section{Limitation and social impact}
\label{sec:conclusion}
\subsection*{Limitation}
While we have made a progressive step in the task of ``Anything Unsegmentable'', successfully creating an attack that is effective and transferable across several models trained under the Segment Anything task, we found it challenging when evaluating our attack on Segment Everything Everywhere Models (SEEM) \cite{zou2023segment}. The reason behind this lack of transferability may stem from fundamental differences in training data and tasks: SEEM is trained on COCO2017 \cite{lin2014microsoft} with panoptic segmentation annotations. Consequently, the feature space of the SEEM model inherently contains rich information about semantic labels, which is significantly different with SAM family. We believe that this divergence in feature space is the primary reason our attack did not transfer successfully.

However, we are optimistic about the potential for improvement. By introducing additional loss term that targets the category feature space, we anticipate the development of new and more powerful adversarial attacks capable of simultaneously compromising SAM, SEEM, and even more promptable segmentation models.

\subsection*{Social Impact}
Our primary goal is to protect the personal digital content from potential copyright infringement and privacy breaches. We envision users employing our approach to preprocess their digital assets before uploading them to public websites, thereby reducing the risk of misuse or theft of their photos and digital creations.

An alternative approach, instead of incorporating adversarial attacks, could involve implementing protective measures directly within the segmentation models themselves. For instance, model publishers might consider adopting a consensus not to perform valid segmentations on protected data. However, establishing and enforcing such a consensus is a complex challenge. Moreover, addressing the issue of models that have already been downloaded and deployed by potential adversaries presents its own set of difficulties.

{
    \small
    \bibliographystyle{ieeenat_fullname}
    \bibliography{main}
}


\end{document}

%% file: preamble.tex
\usepackage[dvipsnames]{xcolor}
\usepackage{algorithm}
\usepackage{algpseudocode}
\usepackage{diagbox}
\usepackage{multirow}

\newcommand{\setParShrink}{\setlength {\parskip} {-0.3cm} }
\newcommand{\setParRecover}{\setlength {\parskip} {0pt}}